\documentclass{article}

\usepackage[preprint]{corl_2026} % Uncomment for pre-prints (e.g., arxiv); This is like ``final'', but will remove the CORL footnote.
\usepackage{algorithm}
\usepackage{algpseudocode}
\usepackage{subcaption}
\usepackage{newfloat}
\usepackage{adjustbox}
\usepackage{listings}
\usepackage[table]{xcolor}
\usepackage{tikz}
\usepackage{mathtools} % for \coloneqq
\usepackage{amsfonts} % 
\usepackage{cleveref}
\usepackage{enumitem}
\Crefname{figure}{Fig.}{Figs.}
\crefname{figure}{Fig.}{Figs.}
\usepackage{wrapfig}
\usepackage{subcaption}
\usepackage{booktabs}
\usepackage{tabularx}

\usepackage[most]{tcolorbox}
\usepackage{parskip} % For spacing between paragraphs
\usetikzlibrary{arrows.meta, positioning, shapes.multipart, matrix, fit}

\usetikzlibrary{arrows, automata, positioning}
\tikzset{auto, >=stealth}
\tikzset{every edge/.append style={shorten >= 1pt}}
\tikzset{
    main node/.style={circle,draw,minimum size=1cm,inner sep=0pt},
}

\DeclareCaptionStyle{ruled}{labelfont=normalfont,labelsep=colon,strut=off} % DO NOT CHANGE THIS
\floatstyle{ruled}

\usepackage{listings}
\definecolor{codegreen}{rgb}{0,0.6,0}
\definecolor{codegray}{rgb}{0.5,0.5,0.5}
\definecolor{codepurple}{rgb}{0.58,0,0.82}
\definecolor{backcolour}{rgb}{0.95,0.95,0.92}

\lstdefinestyle{code}{
    backgroundcolor=\color{backcolour},   
    commentstyle=\color{codegreen},
    keywordstyle=\color{magenta},
    numberstyle=\tiny\color{codegray},
    stringstyle=\color{codepurple},
    basicstyle=\ttfamily\footnotesize,
    breakatwhitespace=false,         
    breaklines=true,                 
    captionpos=b,                    
    keepspaces=true,                 
    numbers=left,                    
    numbersep=5pt,                  
    showspaces=false,                
    showstringspaces=false,
    showtabs=false,                  
    tabsize=2,
    moredelim=**[is][\color{blue}]{@}{@}
}

\lstdefinestyle{base}{
  language=tcl,
  emptylines=1,
  breaklines=true,
  basicstyle=\ttfamily\color{blue},
  moredelim=**[is][\color{purple}]{@}{@},
}

\usepackage{amsthm}
\usepackage{graphicx, array, blindtext}
\usepackage{dblfloatfix}
\usepackage{placeins}
\usepackage{float}
\theoremstyle{definition}

\usepackage{xspace}
\newcommand{\ours}{\texttt{A4D}\xspace}

\usepackage{colortbl}

\usepackage{listings}

\lstdefinestyle{pythonprompt}{
  language=Python,
  basicstyle=\ttfamily\scriptsize,
  breaklines=true,
  columns=fullflexible,
  keepspaces=true,
  showstringspaces=false,
  tabsize=2,
  frame=single,
  framerule=0.4pt,
  aboveskip=0.8\baselineskip,
  belowskip=0.8\baselineskip,
}

\title{What Objects Enable, Not What They Are: Functional Latent Spaces for Affordance Reasoning
}

\author{
  Rohan Siva$^{*1}$ \, Neel P. Bhatt$^{*1,2}$ \, Yunhao Yang$^{*1,2}$ \, Seoyoung Lee$^1$ \, Nishant Gadde$^1$\\
  \textbf{Christian Ellis}$^{1,2}$ \, \textbf{Alvaro Velasquez$^{2,3}$} \, \textbf{Zhangyang Wang$^1$} \, \textbf{Ufuk Topcu$^{1,2}$}\\
  $^1$The University of Texas at Austin \, $^2$Neurosymbolic Intelligence\\
  $^3$University of Colorado Boulder \, $^*$Equal Contribution\\
  \texttt{\{rohansiva,npbhatt,yunhaoyang234,seoyounglee,nishantg\}@utexas.edu}\\
  \texttt{\{chrisitan.ellis@austin.,atlaswang@,utpocu@\}utexas.edu}\\
  \texttt{alvaro.velasquez@colorado.edu}
}

\begin{document}
\maketitle

%===============================================================================

\begin{abstract}

    Existing robot planning systems rely on appearance-based reasoning, where visual observations are encoded into latent spaces organized around object appearances (e.g., recognizing a ``cart'' based on how it looks).
    However, planning requires reasoning about task-relevant functionalities of objects (e.g., whether an object is ``movable''), which appearance-based latent spaces do not capture.
    As a result, existing approaches struggle to generalize to novel robot-object interactions.
    We address this limited generalizability through affordance reasoning, enabling planning based on task-relevant object functionalities instead of appearance alone.
    We introduce \ours, which maps visual observations into a shared latent space structured around affordances (e.g., ``movable").
    By projecting visual observations into this functional latent space and measuring their proximity to affordances, \ours \emph{infers} functionalities relevant to the observed object.
    Furthermore, we introduce an affordance \emph{discovery} mechanism that expands the latent space to handle unseen scenarios where existing affordances are insufficient.
    \ours uses proximity in the functional latent space to quantify uncertainty in affordance inference and selectively triggers affordance discovery.
    We evaluate \ours across several planning tasks involving diverse and unseen affordances. \ours achieves 94\% inference accuracy on existing affordances outperforming state-of-the-art approaches by over 15\% points, improves new-affordance inference accuracy from $\sim$70\% to over 90\% with fewer than $\sim$10\% of the original training data, and enables 100x faster inference. Code, videos, and data available at: \href{https://A4Dance-reasoning.github.io/}{https://A4Dance-reasoning.github.io}.
    
\end{abstract}


\section{Introduction} 
Recent advances in robot planning primarily rely on latent spaces organized around object appearance~\cite{radford2021learning, hassanin2018visualaffordance, affordancenet}.
While effective for recognizing what an object is, appearance-based reasoning often fails to capture how a robot can use the object for a task. For example, recognizing a ``cart'' does not directly indicate whether it is ``movable'' or ``supportable.''
As a result, these systems struggle to generalize to novel robot-object interactions, since visually different objects may support the same functionality while visually similar objects may not.

Planning instead requires reasoning about \emph{affordances}, task-relevant object functionalities defined by the relationship between an object's physical properties and the robot's capabilities~\cite{gibson1979ecological, norman1988psychology}. 
Unlike object categories, affordances describe how an object can be used in a plan, such as whether it is movable, liftable, or supportable.
Affordance-based reasoning therefore provides a more generalizable foundation for planning than appearance alone.
Recent vision-language and interaction-based methods have begun to connect semantic representations with robot affordances, enabling language-grounded planning, open-world manipulation, and spatial affordance prediction~\cite{radford2021learning,ahn2022saycan,tang2024kalie,robopoint2024,belkhale2022plato}.
However, existing approaches often rely on task-specific affordance predictors, robot interaction data, or expensive VLM inference, making it difficult to support fine-grained, uncertainty-aware affordance reasoning for real-time planning.

To enable generalizable and efficient affordance-based reasoning, we introduce \ours, a framework that maps visual observations into a shared \emph{functional latent space} structured around affordances. By reasoning in this latent space, \ours infers task-relevant functionalities directly from visual observations while remaining robust to novel objects and interactions. Furthermore, \ours quantifies uncertainty in affordance inference and selectively expands its affordance memory through affordance discovery when existing affordances are insufficient.

We evaluate \ours across several planning tasks involving diverse and unseen affordances. \ours achieves 94\% inference accuracy on existing affordances, outperforming state-of-the-art approaches by over 20 percentage points, improves new-affordance inference accuracy from $\sim$70\% to over 90\% with fewer than 16 labeled examples, and enables 100$\times$ faster inference for real-time planning. 
We summarize our primary contributions below:

% We summarize our primary contributions as follows:
% (1) \textbf{Functional Latent Spaces for Affordance Reasoning}: We introduce \ours, which maps visual observations into a shared functional latent space structured around affordances, enabling planning based on task-relevant object functionalities instead of appearance alone.\vspace{-0.2em}
% (2) \textbf{Uncertainty-Aware Affordance Discovery}: We introduce an affordance discovery mechanism that uses proximity in the functional latent space to quantify uncertainty in affordance inference and selectively expands the affordance space when existing affordances are insufficient for unseen scenarios.\vspace{-0.2em}
% (3) \textbf{Accurate and Efficient Affordance Inference}: We demonstrate accurate and efficient affordance inference across diverse planning tasks, achieving 94\% inference accuracy on existing affordances, improving new affordance inference accuracy from $\sim$70\% to over 90\% with fewer than 10\% of the original training data, and enabling 100$\times$ faster inference.\vspace{-0.2em}

\begin{figure}[t]
    \centering
    \includegraphics[width=\linewidth]{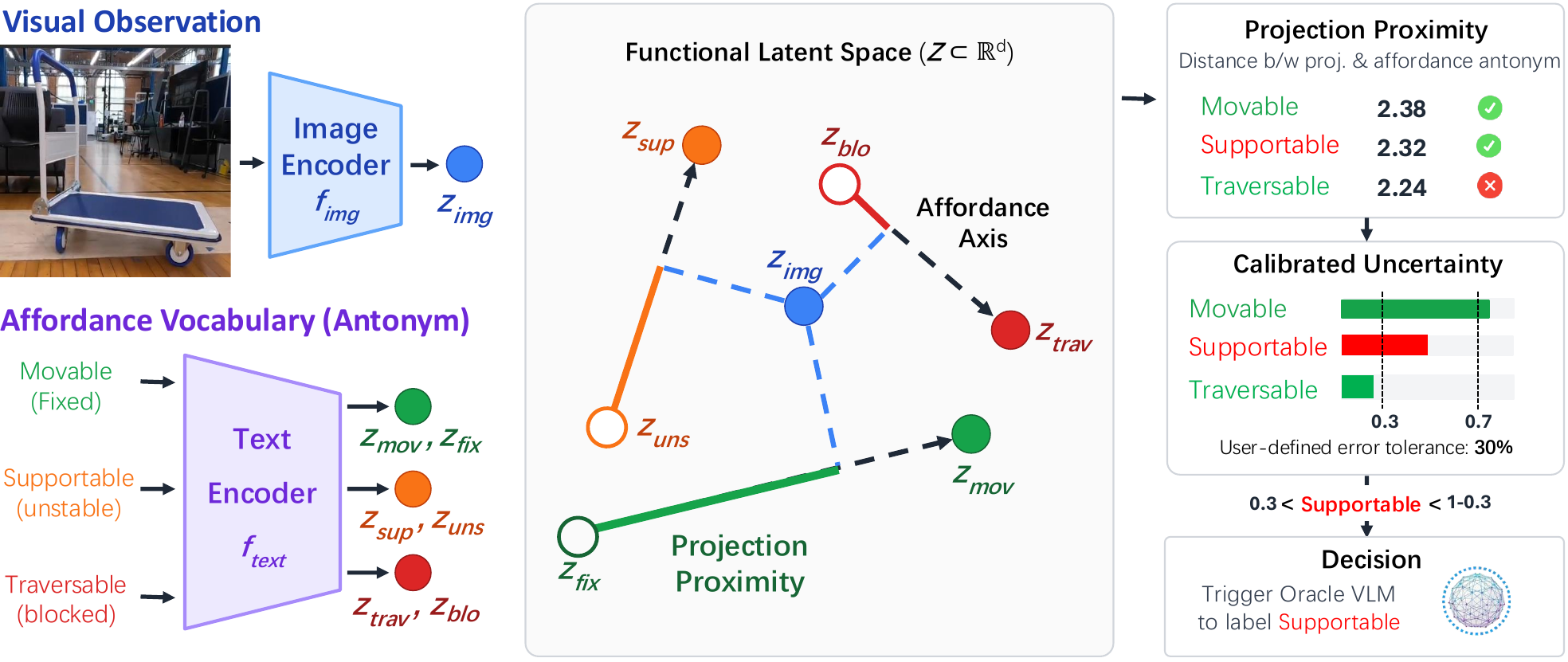}
    \caption{\ours maps visual observations and affordances to a functional latent space.}
    \label{fig:framework}
\end{figure}

\begin{enumerate}[leftmargin=*]
    
    \item \textbf{Functional Latent Spaces for Affordance Reasoning}: We introduce \ours, which maps visual observations into a shared functional latent space structured around affordances, enabling planning based on task-relevant object functionalities instead of appearance alone.\vspace{-0.2em}

    \item \textbf{Accurate and Efficient Affordance Inference}: We demonstrate accurate and efficient affordance inference across diverse planning tasks, achieving 94\% inference accuracy on existing affordances, improving new affordance inference accuracy from $\sim$70\% to over 90\% with fewer than 10\% of the original training data, and enabling 100$\times$ faster inference.\vspace{-0.2em}

    \item \textbf{Uncertainty-Aware Affordance Discovery}: We introduce an affordance discovery mechanism that uses proximity in the functional latent space to quantify uncertainty in affordance inference and selectively expands the space when existing affordances are insufficient for unseen scenarios.\vspace{-0.2em}

\end{enumerate}
\section{Related Work}
\label{sec:related}
Existing work on visual affordances spans appearance-based reasoning,
interaction-based affordance learning, and methods that utilize functional
latent spaces structured with affordances~\cite{hassanin2018visualaffordance,
chuang2017learning2act}. Appearance-based approaches identify affordances in images and videos by labeling regions suitable for actions (grasping, placing, traversing, etc.). Usually, a couple of affordances are specific to object categories or tasks and struggle to generalize beyond their training distribution~\cite{hassanin2018visualaffordance,zeng2019learningvisualaffordances,huang2023affordancesvideos,locate2023,iag2023,afformer2023}. Interaction-based systems like AffordanceNet,
PLATO, SayCan, and ImagineThat learn affordances from robot experience 
through value functions or explicit affordance predictors. These approaches are
usually domain-specific, that are tied to particular robot platforms or skill libraries,
and costly to adapt to new tasks~\cite{affordancenet,belkhale2022plato,
ahn2022saycan,wu2021imaginethat,bahl2023hrp}. Recent work explores vision--language models
and functional latent spaces to move beyond closed-world assumptions and enable
open-vocabulary affordance reasoning, but off-the-shelf models tend to produce
coarse affordance outputs and require high inference time and computing, limiting
their use for a fine-grained, real-time robot
planning~\cite{radford2021learning,robopoint2024,birr2024autogptp,
luo2022learningaffordance}. In contrast, our framework structures a pretrained
vision-language embedding space (CLIP~\cite{radford2021learning}) into a
functional latent space defined by affordance--synonym pairs, supports
uncertainty-aware affordance inference, and enables explicit discovery of new
affordances without retraining, providing a practical bridge between
open-vocabulary affordance reasoning and real-time robot
planning~\cite{tang2024kalie,mazzaglia2024informationaffordance}. A more
detailed survey and discussion of previous work, including additional
appearance-based, interaction-based, and functional latent space methods, is
provided in Appendix~\ref{app:extended_related}.
\section{Problem Formulation}

% Given an image $x \in \mathcal{X}$, we obtain a visual embedding 
% $z_{\mathrm{img}} = f_{\mathrm{img}}(x) \in \mathcal{Z}$, where 
% $f_{\mathrm{img}}$ and $f_{\mathrm{text}}$ map images and text into a shared latent space 
% $\mathcal{Z} \subset \mathbb{R}^d$.

% Each affordance and its antonym $\{a, \bar{a}\} \subset \mathcal{A}$ are defined by a pair of textual descriptions 
% $\{t_a, t_{\bar{a}}\}$. The line joining their embeddings in the latent space define an \emph{affordance axis}: $u_a = f_{\mathrm{text}}(t_a) - f_{\mathrm{text}}(t_{\bar{a}})$. We infer affordance $a$ by projecting the image embedding onto this axis and computing the projection coordinate:
% \[
% \lambda_a(z_{\mathrm{img}}) = \frac{z_{\mathrm{img}}^\top u_a}{\|u_a\|}.
% \]
% The goal is to predict affordance labels $\hat{y}_a \in \{0,1\}$ for all 
% $a \in \mathcal{A}$ from $\lambda_a(z_{\mathrm{img}})$, without task-specific 
% retraining, while generalizing to unseen objects and supporting the addition 
% of new affordances at test time.

We consider a setting where affordances are not directly observable and must be inferred from visual observations, and where the set of relevant affordances may evolve as new tasks and environments are encountered. Given an image $x \in \mathcal{X}$, the goal is to infer binary affordance labels $\hat{y}_a \in \{0,1\}$ for each affordance $a \in \mathcal{A}$, indicating whether the object in the image supports a given task-relevant functionality (e.g., movable, liftable, supportable). 
The objective is to learn a model that predicts affordances from $x$ without task-specific retraining, while generalizing to unseen objects and supporting the addition of new affordances at test time.

\section{Methodology}

We describe the affordance reasoning process in this section, consisting of affordance inference in a functional latent space, uncertainty quantification, affordance discovery, and affordance labeling as illustrated in \Cref{fig:framework2}. We define formal notation for all variables used in this section in Appendix \ref{app:notate_table}.

\begin{figure}[t]
    \centering
    \includegraphics[width=\linewidth]{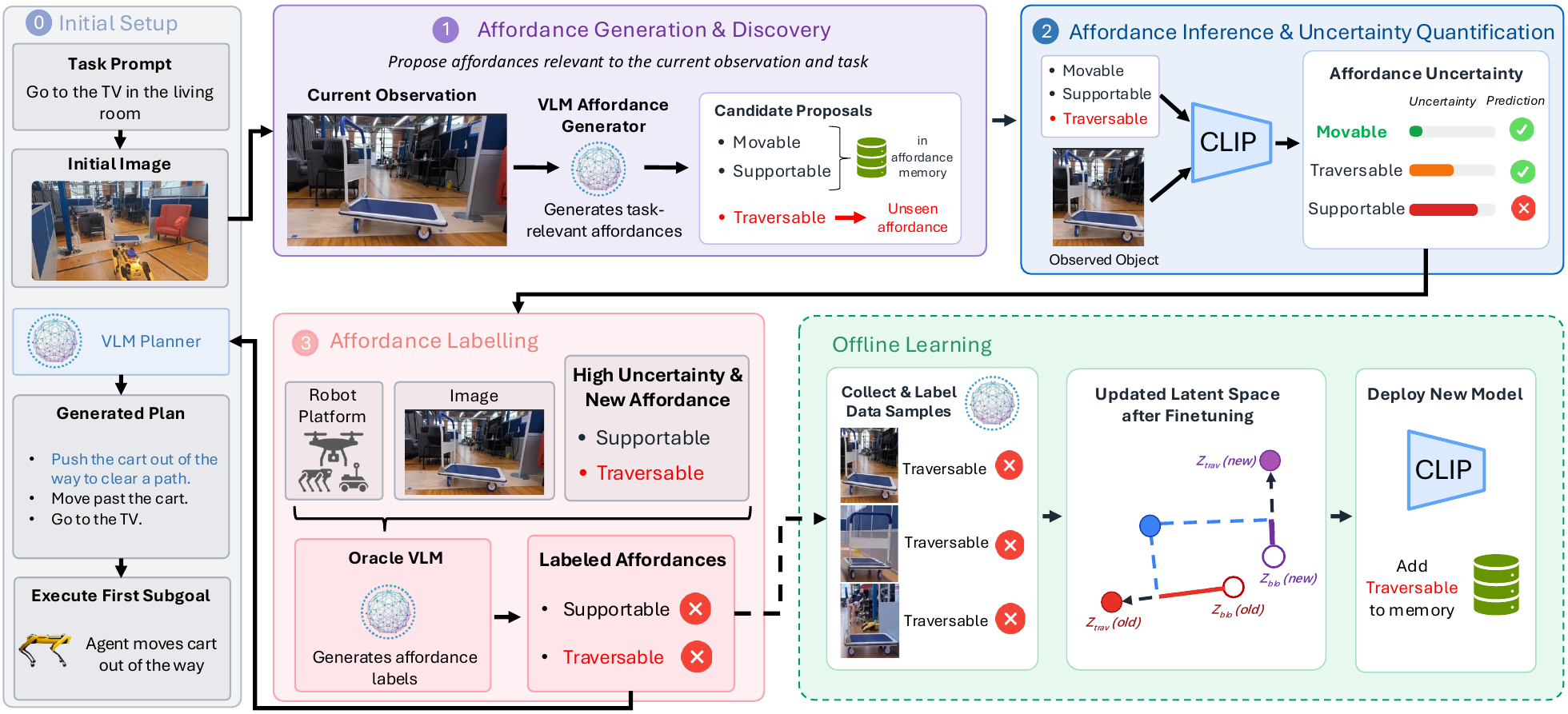}
    \caption{Overview of \ours: consists of affordance generation, discovery, inference, and labeling.}
    \label{fig:framework2}
\end{figure}

\subsection{Affordance Inference}
\label{sec:aff_inf_semantic_affordance_axes}
We infer object affordances by leveraging the semantic structure of a pretrained vision-language embedding space (CLIP \cite{radford2021learning}), where both images and text are mapped into a shared functional latent space. We illustrate this functional latent space in \Cref{fig:framework}.

As shown in \Cref{fig:framework}, we construct an \emph{affordance axis} by connecting the text embeddings of an affordance and its antonym. This axis captures the semantic contrast between the presence and absence of the affordance. Given an image, we project its embedding onto the corresponding affordance axis and determine its relative position along this direction. To further improve alignment between visual observations and affordances, we fine-tune the functional latent space on a small set of labeled affordance-image pairs.

\subsubsection{Affordance Axis in the Functional Latent Space}

Let $\mathcal{Z} \subset \mathbb{R}^d$ denote a shared latent space induced by a pretrained vision--language model (e.g., CLIP), consisting of an image encoder 
$f_{\mathrm{img}}: \mathcal{X} \to \mathcal{Z}$ and a text encoder 
$f_{\mathrm{text}}: \mathcal{T} \to \mathcal{Z}$.
Both encoders map inputs into a functional latent space where semantic similarity is captured by cosine similarity.

Given an affordance $a$, we define two textual descriptions:
(i) a positive description $t_a \in \mathcal{T}$ characterizing the presence of the affordance, and 
(ii) an antonym description $t_{\bar{a}} \in \mathcal{T}$ characterizing its absence.
Their embeddings are given by: $
z_a = f_{\mathrm{text}}(t_a)$ and $z_{\bar{a}} = f_{\mathrm{text}}(t_{\bar{a}})$.

We define the \emph{semantic affordance axis} as the one-dimensional affine subspace connecting these two embeddings. The corresponding direction vector is: $u_a = z_a - z_{\bar{a}}$.

By construction, the positive direction of $u_a$ points from the antonym toward the affordance, and its magnitude encodes the semantic separation between the two. This construction assumes that the pretrained latent space encodes meaningful semantic relations such that the difference vector $u_{a}$ captures the principal direction distinguishing the presence and absence of the affordance.

\paragraph{Learning a Functional Latent Space.}
While the semantic axis is defined purely from text, the alignment between visual embeddings and this axis can be further improved using a small set of labeled images. 
Let $
\mathcal{D}_a = \{(x_i, y_i)\}_{i=1}^n, \, y_i \in \{0,1\},
$
denote a dataset where $y_i = 1$ indicates the presence of affordance $a$.

We fine-tune the encoders $f_{\mathrm{img}}$ and $f_{\mathrm{text}}$ using a contrastive objective that encourages images to align with the correct textual concept. 
Specifically, for each sample $(x_i, y_i)$, we define the target text embedding
\[
z_i^{+} =
\begin{cases}
z_a, & y_i = 1, \\
z_{\bar{a}}, & y_i = 0,
\end{cases}
\qquad
z_i^{-} =
\begin{cases}
z_{\bar{a}}, & y_i = 1, \\
z_a, & y_i = 0.
\end{cases}
\]

Let $z_i = f_{\mathrm{img}}(x_i)$ denote the image embedding. 
We adopt a contrastive loss of the form
\[
\mathcal{L}_i = - \log \frac{\exp(\mathrm{sim}(z_i, z_i^{+}) / \tau)}
{\exp(\mathrm{sim}(z_i, z_i^{+}) / \tau) + \exp(\mathrm{sim}(z_i, z_i^{-}) / \tau)},
\]

where $\mathrm{sim}(\cdot,\cdot)$ denotes cosine similarity and $\tau > 0$ is a temperature parameter.

This objective encourages embeddings of images with affordance $a$ to align with $z_a$ and repel from $z_{\bar{a}}$, and vice versa for negative samples. As a result, the visual embeddings are reshaped such that their projections along the semantic affordance axis become more discriminative, while preserving the global structure of the pretrained latent space. The final affordance axis $u_a$ is defined by the (optionally fine-tuned) text embeddings, and is shared across all images for downstream inference.

\subsubsection{Affordance Inference using Projection Proximity}

Given an image $x$, we obtain its embedding $z_{img} = f_{\mathrm{img}}(x) \in \mathcal{Z}$. We project $z_{img}$ onto the semantic affordance axis defined by $(z_{\bar{a}}, z_a)$.
The scalar projection coordinate is given by
\[
\lambda_a(z_{img}) = \langle z_{img} - z_{\bar{a}},\, u_a \rangle \ /\  \|u_a\|^2,
\]
where $\langle , \rangle$ is the standard inner product in $\mathbb{R}^d$, and $\|\cdot\|$ denotes the Euclidean norm.

This coordinate represents the relative position of $z_{img}$ along the axis, where
$\lambda_a(z_{img}) = 0$ corresponds to the antonym affordance and 
$\lambda_a(z_{img}) = 1$ corresponds to the affordance itself.

We classify the affordance based on the midpoint of the axis:
\[
\hat{y}_a =
\begin{cases}
1, & \lambda_a(z_{img}) > \tfrac{1}{2}, \\
0, & \lambda_a(z_{img}) \le \tfrac{1}{2}.
\end{cases}
\]

This formulation yields a simple geometric decision rule, where affordance prediction reduces to determining whether the projected embedding lies closer to the affordance or its antonym in the functional latent space.

\subsubsection{Uncertainty Quantification}

For each affordance $a$, we clip the projection coordinate $\lambda_a(z_{img})$ to obtain a normalized axis score $r_a(z_{img}) = \mathrm{clip}\!\left(\lambda_a(z_{img}),0,1\right)$. By construction, $r_a(z_{img})=0$ corresponds to the antonym embedding and $r_a(z_{img})=1$ corresponds to the affordance embedding.
We interpret $r_a(z_{img})$ as an uncalibrated score for the presence of affordance $a$. Let $y \in \{0,1\}$ denote whether affordance is actually present in image $x$. We calibrate the axis score by estimating 
\[
\eta(x, a) = \Pr\big(y=1 \mid r_a(z_{img})\big) = \Pr\big(y=1 \ | \ r_a(f_{\mathrm{img}}(x))\big).
\]
Under the assumption that the affine axis coordinate is order-preserving with respect to affordance presence, $\eta(x, a)$ is monotone in $r_a(z_{img})$. We therefore estimate this mapping using isotonic regression on a held-out calibration set
$
\mathcal{D}_{\mathrm{cal}} = \{(x_i,y_i)\}_{i=1}^{n},
$
which is disjoint from the training data.
The calibrated output $\eta(x, a)$ estimates the probability that affordance $a$ actually exists in the image.

Given a user-specified prediction error tolerance threshold $\tau \in (0,0.5)$, we consider the prediction \textbf{uncertain} when $\tau < \eta(x,a) < 1-\tau$.
Otherwise, $\eta(x,a)\leq \tau$ means that affordance $a$ is confidently absent, whereas $\eta(x,a)\geq 1-\tau$ indicates high confidence that affordance $a$ is present.

\subsection{Affordance Discovery}
\label{sec:affordance_discovery}

A fixed affordance memory, $\mathcal{A}_{\mathrm{base}}$, limits generalization when robots encounter objects with unseen affordances. \ours{} addresses this by selectively expanding the affordance memory when existing affordances are insufficient. For each object image $x_i$ and task context $\tau_i$, we query a vision-language (VLM) with task-centric instructions to generate a candidate affordance set $\mathcal{A}_i$ (eg, \texttt{\{Movable, Supportable, Traversable\}}) consisting of affordances that are relevant to the task. Candidates may either correspond to existing affordances in the base memory or introduce new affordances: $\mathcal{A}_i
=
\mathcal{A}_{i}^{\mathrm{base}}
\cup
\mathcal{A}_{i}^{\mathrm{new}},
$
where $\mathcal{A}_{i}^{\mathrm{base}} \subseteq \mathcal{A}_{\mathrm{base}}$ are reused affordances (eg, \texttt{\{Movable, Supportable\}}) and $\mathcal{A}_{i}^{\mathrm{new}}$ are newly proposed affordances (eg, \texttt{\{Traversable\}}). Full discovery prompts are provided in Appendix \ref{app:prompts_aff_gen_labeling}. Candidate affordances are compared against the current base memory bank of new affordances used during training.

\subsection{Affordance Labeling}
\label{sec:affordance_labeling}

The discovery stage proposes candidate affordances, but these candidates still require labels before they can be used for training, evaluation, or axis construction. Moreover, existing affordance may yield very high uncertainty scores. Either of these conditions triggers affordance labelling. We label each proposed pair $(x_i, a_{ij})$, where $a_{ij} \in \mathcal{A}_i$, using a second oracle VLM query. The model receives only the cropped object image $x_i$ and a single affordance $a_{ij}$ and predicts a binary label $\tilde{y}_{ij} \in \{0,1\}$ indicating whether the visible object clearly supports that affordance. The prompts used for labeling can be found in \Cref{app:prompts_aff_gen_labeling}. 
The resulting labeled set is
$\tilde{\mathcal{D}}_{\mathrm{disc}}
    =
    \{(x_i, a_{ij}, \tilde{y}_{ij})
    \mid a_{ij} \in \mathcal{A}_i^{\mathrm{prop}}\}.
$
The labeled set can be used to construct new affordance axes or support incremental affordance learning. This two-stage design avoids exhaustively querying every affordance for every object: \ours{} first proposes a small canonicalized and validated candidate set, then labels only those candidates. After each discovery batch, the batch-level proposal set $\mathcal{A}^{\mathrm{prop}}=\bigcup_i \mathcal{A}_i^{\mathrm{prop}}$ becomes the base memory $\mathcal{A}_{\mathrm{base}}$ for the next discovery iteration.

\section{Experimental Evaluation}
\label{sec:exp}
Our experiments evaluate accurate and efficient affordance inference on known affordances (§\ref{sec:exp-accuracy}), sample-efficient acquisition of new affordances (§\ref{sec:exp-discovery}), and uncertainty-guided VLM fallback (§\ref{sec:exp-uncertainty}). We then demonstrate these components in an end-to-end planning task where uncertainty triggers affordance discovery (§\ref{sec:exp-endtoend}).

% Across all experiments, pretrained vision-language models lack explicit affordance grounding, but can be effectively adapted with relatively small amounts of labeled data. Our fine-tuned CLIP model achieves high accuracy, strong sample efficiency, and low inference latency, making it well-suited for integration into real-time robotic systems.
% Supporting analyses appear in the appendix: label-noise robustness up to $\sim$20\% noise from VLM-generated supervision (Appendix~\cref{app:label-noise}), stability--plasticity behavior under sequential affordance additions (Appendix~\cref{app:incremental}), and extended uncertainty-triggered fallback results (Appendix~\cref{app:uncertainty-fallback}). Together, these establish that the headline numbers are not artifacts of clean supervision and that adding new affordances does not destabilize previously learned ones.

\subsection{Accurate and Real-Time Inference on Known Affordances}
\label{sec:exp-accuracy}

\Cref{tab: baseline_performance_AE} reports affordance inference accuracy and latency across five evaluation settings (A--D), across seen and unseen object classes and affordances. Setting A corresponds to seen affordances and seen object classes.
\ours{} attains \textbf{94.2\% accuracy on setting A}, outperforming the strongest VLM baseline (GPT-5.4 medium, 78.9\%) by 15.3 points. Zero-shot vision--language baselines remain near chance (CLIP 48\%, BLIP 45\%), indicating that pretrained semantic similarity alone does not recover the functional distinctions required for affordance reasoning. A per-affordance breakdown in Appendix~\ref{app:per-affordance} shows accuracy $\geq$0.981 across all eight base affordances.

\ours{} generalizes well to unseen object classes, but performance drops with unseen affordances. With unseen object classes and seen affordances (setting C), \ours{} drops only slightly to 86.3\%, showing that the learned affordance axes transfer across unseen objects. In contrast, performance drops when affordances are unseen (settings B, D), where \ours{} falls to 59--61\%. To mitigate this gap, we introduce a discovery mechanism in \S\ref{sec:exp-discovery} that expands the seen affordance memory and an uncertainty-guided fallback in \S\ref{sec:exp-uncertainty} to trigger high-performance VLM labeling.

\paragraph{Real-time inference latency.}
\ours{} infers a single affordance in \textbf{22\,ms}, over \textbf{100$\times$ faster} than prompted GPT-5 baselines. VLM-based methods incur high inference latency, making continuous affordance reasoning impractical for closed-loop planning. In contrast, \ours{} provides low-latency affordance inference while preserving strong in-distribution and unseen-object performance, enabling real-time affordance reasoning.

\begin{table}[!htbp]
\centering
\caption{Accuracy and per-affordance inference latency across five evaluation settings. Settings vary object-class and affordance novelty: \textbf{S}=seen, \textbf{U}=unseen.}
\label{tab: baseline_performance_AE}
\begin{tabular}{lccccc}
\hline
Model & A (S/S) & B (S/U) & C (U/S) & D (U/U) & Latency (ms) \\
\hline

CLIP zero-shot         & 0.481 & 0.308 & 0.594 & 0.400 & 21 \\
SigLIP zero-shot       & 0.625 & 0.369 & 0.581 & 0.415 & 20 \\
BLIP zero-shot         & 0.452 & 0.708 & 0.425 & 0.690 & 84 \\
ViLT-B32 (VQA)         & 0.587 & 0.182 & 0.377 & 0.228 & 9 \\
GPT-5-nano             & 0.538 & 0.823 & 0.738 & 0.820 & 4{,}100 \\
GPT-5-nano CoT         & 0.635 & 0.823 & 0.706 & 0.830 & 20{,}500 \\
GPT-5-mini             & 0.702 & 0.854 & 0.750 & 0.815 & 3{,}900 \\
GPT-5.4 (low)          & 0.731 & 0.839 & 0.681 & 0.815 & 1{,}871 \\
GPT-5.4 (medium)       & 0.789 & 0.854 & 0.794 & 0.825 & 2{,}646 \\
\hline
A4D (CLIP frozen-text)   & 0.875 & 0.508 & \textbf{0.869} & 0.665 & \textbf{22} \\
A4D (CLIP frozen-vision) & \textbf{0.962} & 0.562 & 0.844 & 0.540 & \textbf{22} \\
\textbf{A4D (CLIP unfrozen)}      & 0.942 & 0.608 & 0.863 & 0.605 & \textbf{22} \\
\hline
\end{tabular}

\end{table}

\FloatBarrier

\subsection{Sample-Efficient Acquisition of New Affordances}
\label{sec:exp-discovery}

\begin{figure}[b]
    \centering
    \includegraphics[width=\linewidth]{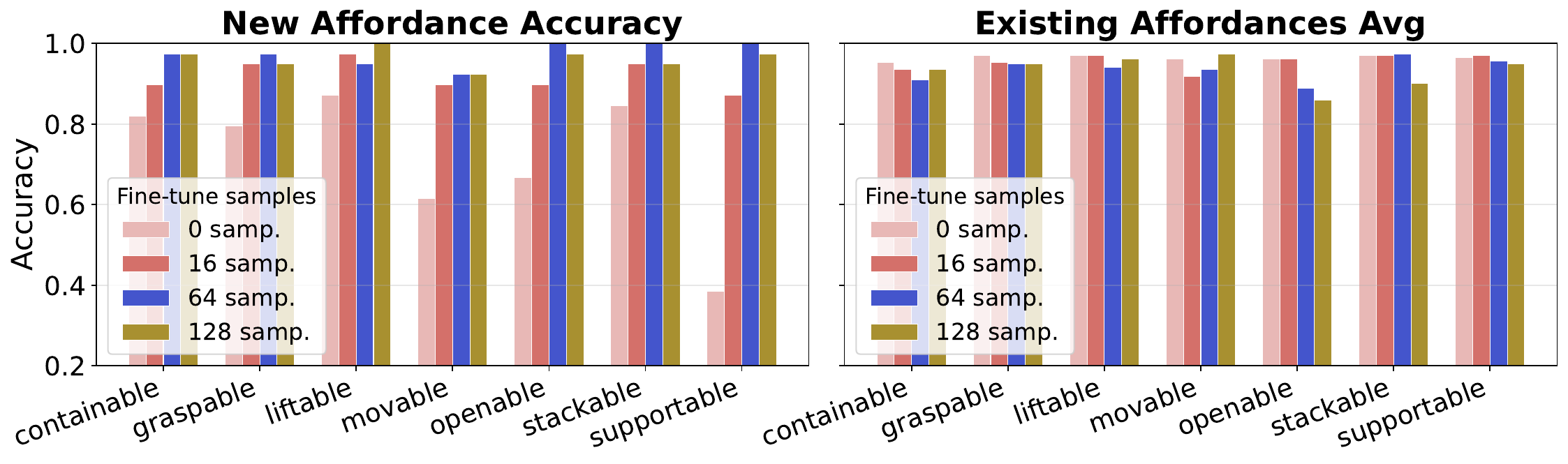}
    \caption{
    Leave-one-out affordance acquisition and existing affordance forgetting. 
    (L) Accuracy of new affordances vs. number of samples. 
    (R) Average accuracy of seed affordances after adding.
    }
    %\textbf{Left}: held-out affordance accuracy after introducing labeled fine-tuning samples. \textbf{Right}: average accuracy on previously learned affordances after adding the new affordance.
    \label{fig:loo-results}
\end{figure}

We now measure how efficiently the discovery mechanism (\S\ref{sec:affordance_discovery}, \S\ref{sec:affordance_labeling}) closes the unseen-affordance performance gap identified in \S\ref{sec:exp-accuracy}.
When the calibrated uncertainty scores indicate that existing affordances are insufficient, a VLM proposes a candidate affordance and labels a small set of examples for offline finetuning.
% The VLM labels are noisy, but projection-based inference tolerates the residual error: accuracy degrades only marginally up to $\sim$15\% label noise, near the empirical GPT-5.4 labeling error.
% We therefore use VLM supervision directly, without an explicit filtering stage.
\cref{fig:loo-results} isolates sample efficiency from label noise by using ground-truth labels for the held-out affordance.
We measure sample efficiency by holding out one of the seven affordances during initial training, then reintroducing it with varying numbers of labeled examples.
% \cref{fig:loo-results} isolates sample efficiency from label noise by measuring it with a leave-one-out (LOO) experiment that uses ground-truth labels for the held-out affordance. 
% We remove each of the seven affordances from training and reintroduce it with varying numbers of labeled examples.

\paragraph{New affordances saturate at 16 labels.}
Average accuracy on the held-out affordance rises from $\sim$70\% with no fine-tuning samples to over 92\% with just 16 labeled examples ($\sim$10\% of the original training data). This low-label performance substantially reduces the amount of VLM supervision required to acquire a new affordance. Fine-tuning with a small number of labeled examples reshapes the functional latent space around the new affordance axis, converting weak zero-shot unseen-affordance performance into reliable affordance inference.

\paragraph{Prior affordances are preserved.}
The right side of \cref{fig:loo-results} measures how adding a new affordance affects previously learned affordance axes. Smaller fine-tuning sets preserve prior affordances more effectively: 16-sample updates cause only minor degradation in existing affordance accuracy, while larger updates (e.g., 128 samples) introduce substantially more forgetting. This suggests that small latent-space updates are sufficient to incorporate new affordances without significantly disrupting previously learned affordance structure. Appendix~\ref{app:incremental} extends this analysis to sequential affordance additions, where repeated large updates lead to catastrophic forgetting while few-shot updates better preserve prior affordances.

\begin{wrapfigure}{r}{0.48\textwidth}
    \centering
    \vspace{-10pt}
    \includegraphics[width=\linewidth]{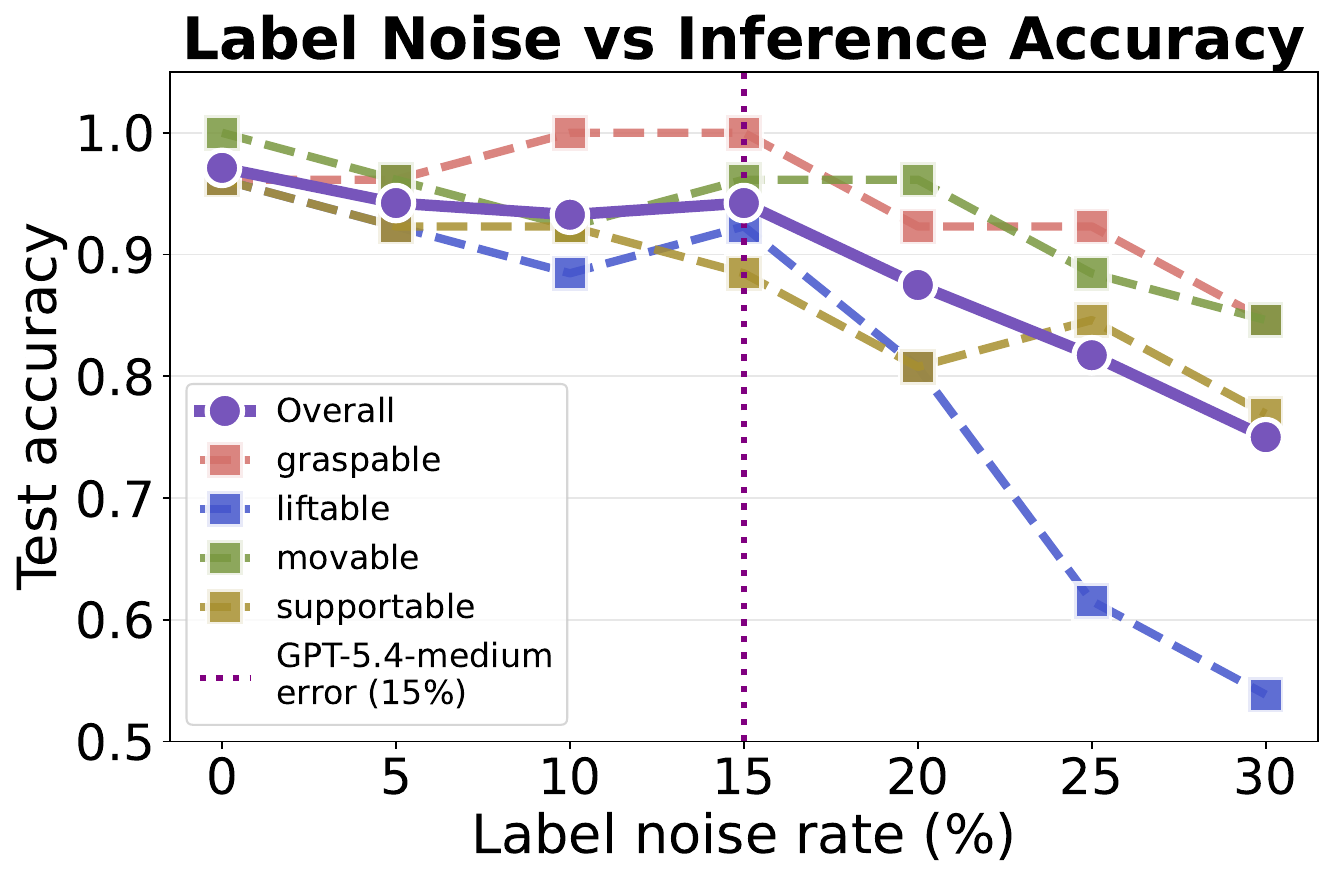}
    \vspace{-8pt}
    \caption{Inference accuracy trend.}
    \label{fig:label-noise}
    \vspace{-10pt}
\end{wrapfigure}

\paragraph{Robustness to VLM label noise.}
Because affordance discovery and labeling rely on VLM supervision (\S\ref{sec:affordance_discovery}, \S\ref{sec:affordance_labeling}), we test how the resulting label noise affects downstream inference.
GPT-5.4's labeling error reaches $\sim$15\%, so we inject controlled label noise during training and measure inference accuracy (\cref{fig:label-noise}).
The fine-tuned CLIP model degrades only marginally from 0\% to 15\% noise, indicating that projection-based inference is robust within the GPT-5.4 error regime.

\subsection{Uncertainty-Guided Fallback}
\label{sec:exp-uncertainty}

\begin{figure}[H]
    \centering
    \includegraphics[width=0.85\linewidth]{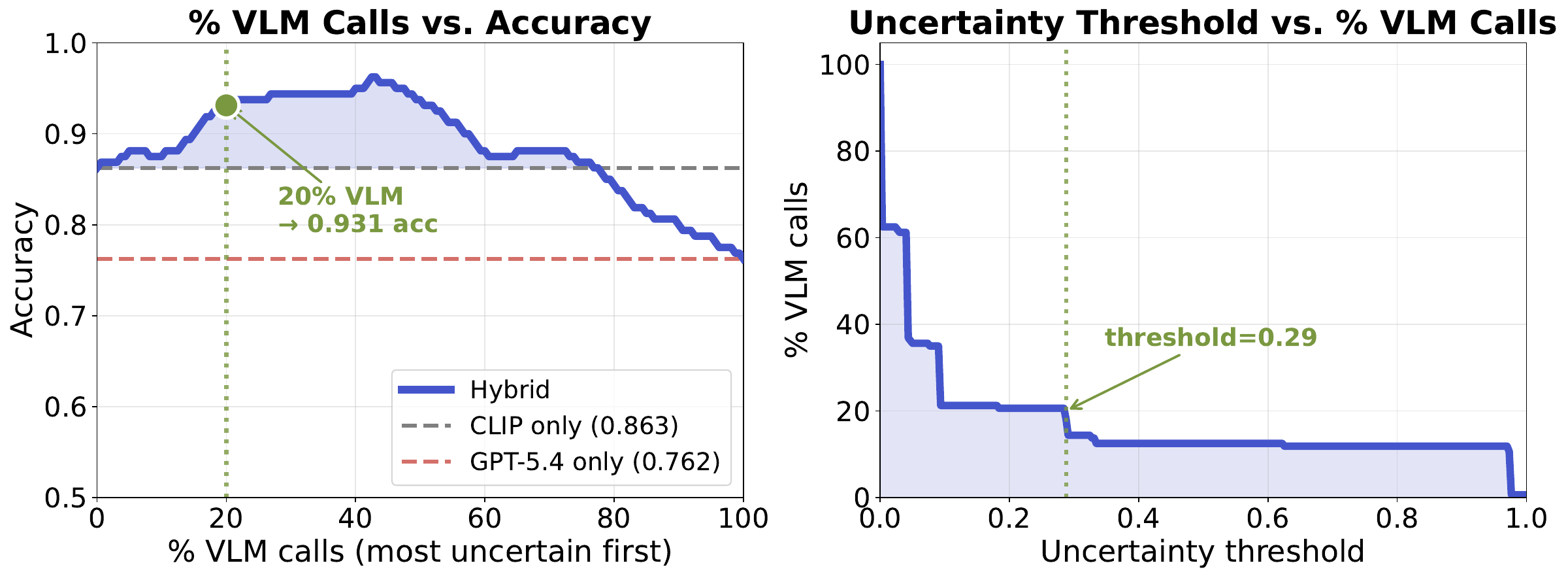}
 \caption{
    Uncertainty-guided VLM trigger: Accuracy vs Query Frequency tradeoff.
    (L) Accuracy vs. fraction of VLM calls. (R) Percentage of VLM calls as the uncertainty threshold is swept.
    }
    \label{fig:unc-accuracy}
\end{figure}

We evaluate whether projection-based uncertainty estimates can effectively determine when additional VLM reasoning is needed. We evaluate on unseen object classes and seen affordances, in which GPT-5.4 serves as an oracle, invoked only when the uncertainty exceeds a threshold. Low-uncertainty predictions are handled directly by \ours{}, while high-uncertainty cases receive additional VLM labeling and reasoning.
\Cref{fig:unc-accuracy} (left) shows that \ours{} achieves 93\% accuracy while invoking VLM fallback for less than 20\% of the highest uncertainty inferences.
\Cref{fig:unc-accuracy} (right) sweeps the uncertainty threshold controlling this query frequency, and \cref{fig:Qualitative_Demos} presents qualitative examples where uncertainty guides object interaction selection.
Overall, uncertainty serves as an effective trigger for selectively targeting likely failure cases and improving affordance inference accuracy with only a small fraction of VLM calls.

\subsection{End-to-End Planning Demonstration}
\label{sec:exp-endtoend}

\begin{figure}[!htbp]
    \centering
    \includegraphics[width=1\linewidth]{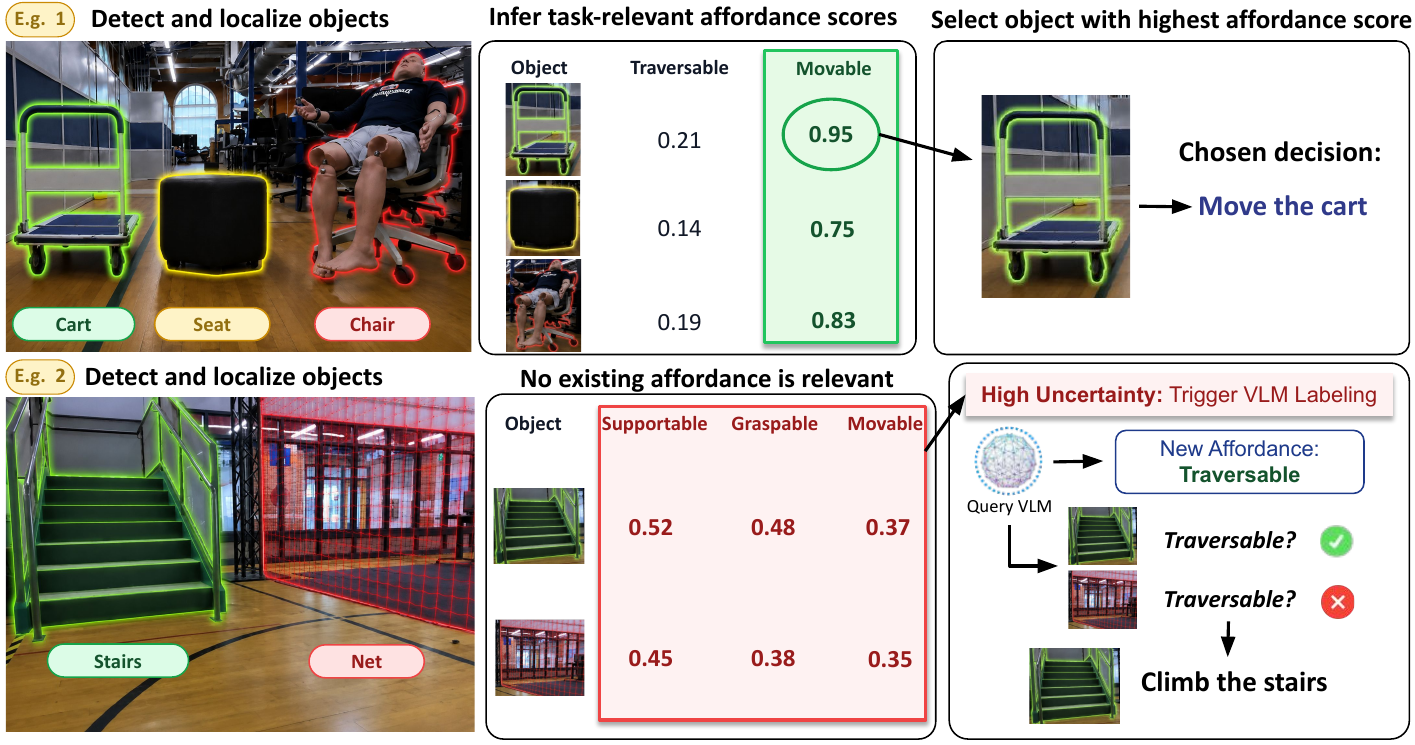}
    \caption{\ours{} in two independent deployment scenarios.
    \textbf{E.g. 1} (``Move the cart'') selects an action based on Uncertainty.
    \textbf{E.g. 2} (``Climb the stairs'') triggers affordance discovery, adding \emph{Traversable}.}
    \label{fig:Qualitative_Demos}
    \vspace{-10pt}
\end{figure}

The preceding subsections evaluated accuracy (C1, \S\ref{sec:exp-accuracy}), discovery efficiency (C2, \S\ref{sec:exp-discovery}), and uncertainty-guided fallback (C3, \S\ref{sec:exp-uncertainty}) independently.
\Cref{fig:Qualitative_Demos} demonstrates these capabilities operating together in a planner across two independent scenes.
In each, \ours{} detects candidate objects, scores them on the current affordance axes, and selects one for the task.

In e.g. 1 (``Move the cart''), the \textit{Traversable} and \textit{Movable} affordances are pre-seeded in the memory.
The planner identifies the cart as the lowest uncertainty movable object and completes the task without VLM fallback, demonstrating uncertainty-based object interaction.

In e.g. 2 (``Climb the stairs''), the base affordances \textit{Supportable}, \textit{Graspable}, and \textit{Movable} all produce uncertain predictions near the decision boundary.
This uncertainty triggers affordance discovery (\S\ref{sec:affordance_discovery}): a VLM proposes a new affordance, ``Traversable'', labels each object along that axis (\S\ref{sec:affordance_labeling}), and selects the stairs.

Together, these examples show that VLM supervision is used only during memory expansion and high-uncertainty labeling, while downstream planning operates directly on the affordance memory.

\section{Conclusion}

We develop \ours, a framework for affordance-based robot reasoning and planning that shifts reasoning from object appearance to task-relevant functionalities. By learning a functional latent space, \ours enables more generalizable reasoning across diverse and unseen robot-object interactions. Furthermore, the framework includes an uncertainty-aware affordance discovery mechanism supports adaptation to previously unseen affordances with minimal labeled data. Experimental results demonstrate strong improvements in accuracy, adaptability, and efficiency, highlighting the promise of affordance-centric latent representations for robust and scalable robot planning.

\section{Limitations and Future Work}
While \ours{} demonstrates strong performance in affordance inference, a few limitations remain: 
\textbf{(1)} \ours{} focuses on affordance reasoning rather than downstream planning performance, and although inferred affordances are passed to a high-level planner, this integration is ad-hoc rather than a principled planner-aware evaluation.
\textbf{(2)} affordance axes are fixed after training, limiting adaptation when inferred affordances disagree with observed interaction outcomes. 
\textbf{(3)} Our uncertainty estimation relies on isotonic calibration, which, despite strong empirical performance, does not provide formal probabilistic guarantees.
Future work will embed \ours{} within a full autonomy stack and explore online adaptation of affordance axes from interaction feedback.
% A natural extension of \ours{} is to learn a correction vector in the embedding space that nudges the axis toward the post-interaction evidence, enabling per-deployment calibration without retraining the backbone.
% Our contribution is not the high level planner, currently there is a handoff to the high level planner that happens ad hoc, later we want to actually test with a planner for performance. 
% We currently use isotonic uncertainty calibration, however since its regression based, it doesnt provide probabalistic guarantees. thus we want to use robust conformal prediction.

\clearpage
\bibliography{refs}  % .bib

@inproceedings{radford2021learning,
  title        = {Learning transferable visual models from natural language supervision},
  author       = {Radford, Alec and Kim, Jong Wook and Hallacy, Chris and Ramesh, Aditya and Goh, Gabriel and Agarwal, Sandhini and Sastry, Girish and Askell, Amanda and Mishkin, Pamela and Clark, Jack and others},
  booktitle    = {International Conference on Machine Learning},
  pages        = {8748--8763},
  year         = {2021},
  organization = {PMLR}
}

@article{hassanin2018visualaffordance,
  title   = {Visual Affordance and Function Understanding: A Survey},
  author  = {Hassanin, Mohammed and Khan, Salman and Tahtali, Murat},
  journal = {arXiv preprint arXiv:1807.06775},
  year    = {2018}
}

@phdthesis{zeng2019learningvisualaffordances,
  title   = {Learning Visual Affordances for Robotic Manipulation},
  author  = {Zeng, Andy},
  school  = {Princeton University},
  address = {Princeton, NJ},
  year    = {2019}
}

@misc{zhu2023visualaffordancelearning,
  title        = {Visual Affordance Learning for Robot Manipulation},
  author       = {Zhu, Yuke},
  howpublished = {Tutorial talk, UT Austin Robot Perception and Learning Lab},
  year         = {2021},
  note         = {Accessed 2026-04-28}
}

@article{affordancenet,
  title   = {AffordanceNet: An End-to-End Deep Learning Approach for Object Affordance Detection},
  author  = {Do, Thanh-Toan and Nguyen, Anh and Reid, Ian},
  journal = {arXiv preprint arXiv:1709.07326},
  year    = {2017}
}

@article{chuang2017learning2act,
  title   = {Learning to Act Properly: Predicting and Explaining Affordances from Images},
  author  = {Chuang, Ching-Yao and Li, Jiaman and Torralba, Antonio and Fidler, Sanja},
  journal = {arXiv preprint arXiv:1712.07576},
  year    = {2017}
}

@article{huang2023affordancesvideos,
  title   = {Affordances from Human Videos as a Versatile Representation for Robotics},
  author  = {Bahl, Shikhar and Mendonca, Russell and Chen, Lili and Jain, Unnat and Pathak, Deepak},
  journal = {arXiv preprint arXiv:2304.08488},
  year    = {2023}
}

@inproceedings{belkhale2022plato,
  title        = {PLATO: Predicting Latent Affordances Through Object-Centric Play},
  author       = {Belkhale, Suneel and Sadigh, Dorsa},
  booktitle    = {Proceedings of the Conference on Robot Learning},
  year         = {2022},
  organization = {PMLR}
}

@inproceedings{wu2021imaginethat,
  title        = {Imagine That! Leveraging Emergent Affordances for 3D Tool Synthesis},
  author       = {Wu, Yizhe and Kasewa, Sudhanshu and Groth, Oliver and Salter, Sasha and Sun, Li and Posner, Ingmar and Gal, Yarin},
  booktitle    = {Proceedings of the IEEE/CVF International Conference on Computer Vision},
  year         = {2021}
}

@article{mazzaglia2024informationaffordance,
  title   = {Information-driven Affordance Discovery for Efficient Robotic Manipulation},
  author  = {Mazzaglia, Pietro and Cohen, Taco and Dijkman, Daniel},
  journal = {arXiv preprint arXiv:2405.03865},
  year    = {2024}
}

@article{birr2024autogptp,
  title   = {AutoGPT+P: Affordance-based Task Planning with Large Language Models},
  author  = {Birr, Timo and Pohl, Christoph and Younes, Abdelrahman and Asfour, Tamim},
  journal = {arXiv preprint arXiv:2402.10778},
  year    = {2024}
}

@article{tang2024kalie,
  title   = {KALIE: Fine-Tuning Vision-Language Models for Open-World Manipulation without Robot Data},
  author  = {Tang, Grace and Rajkumar, Swetha and Zhou, Yifei and Chen, Dian and Zhou, Yilun and Kollar, Thomas and Xiao, Teddy and Pinto, Lerrel and Song, Shuran and Torralba, Antonio and Finn, Chelsea and Agrawal, Pulkit and others},
  journal = {arXiv preprint arXiv:2409.14066},
  year    = {2024}
}

@inproceedings{robopoint2024,
  title        = {RoboPoint: A Vision-Language Model for Spatial Affordance Prediction in Robotics},
  author       = {Yuan, Wentao and Li, Xuesong and others},
  booktitle    = {Proceedings of the 8th Conference on Robot Learning},
  year         = {2025},
  organization = {PMLR}
}

@inproceedings{ahn2022saycan,
  title        = {Do As I Can, Not As I Say: Grounding Language in Robotic Affordances},
  author       = {Ahn, Michael and Brohan, Anthony and Brown, Noah and Chebotar, Yevgen and Cortes, Omar and others},
  booktitle    = {Proceedings of the Conference on Robot Learning},
  year         = {2022},
  organization = {PMLR}
}

@inproceedings{luo2022learningaffordance,
  title     = {Learning Affordance Grounding from Exocentric Images},
  author    = {Luo, Hongchen and Zhai, Wei and Zhang, Jing and Cao, Yang and Tao, Dacheng},
  booktitle = {Proceedings of the IEEE/CVF Conference on Computer Vision and Pattern Recognition},
  year      = {2022}
}

@inproceedings{bahl2023hrp,
  title        = {Human Affordances for Robotic Pre-Training},
  author       = {Bahl, Shikhar and others},
  booktitle    = {Robotics: Science and Systems},
  year         = {2024},
  url          = {https://www.roboticsproceedings.org/rss20/p068.pdf}
}

@inproceedings{locate2023,
  title        = {LOCATE: Localize and Transfer Object Parts for Weakly Supervised Affordance Grounding},
  author       = {Li, G},
  booktitle    = {Proceedings of the IEEE/CVF International Conference on Computer Vision},
  year         = {2023},
  url          = {https://arxiv.org/abs/2303.09665}
}

@inproceedings{iag2023,
  title        = {Grounding 3D Object Affordance from 2D Interactions in Images},
  author       = {Yang, Y},
  booktitle    = {Proceedings of the IEEE/CVF International Conference on Computer Vision},
  year         = {2023},
  url          = {https://arxiv.org/abs/2303.10437}
}

@inproceedings{afformer2023,
  title        = {Affordance Grounding from Demonstration Video to Target Image},
  author       = {Chen, J},
  booktitle    = {Proceedings of the IEEE/CVF International Conference on Computer Vision},
  year         = {2023},
  url          = {https://arxiv.org/abs/2303.14644}
}

@book{gibson1979ecological,
  title     = {The Ecological Approach to Visual Perception},
  author    = {Gibson, James J.},
  year      = {1979},
  publisher = {Houghton Mifflin},
  address   = {Boston, MA}
}

@book{norman1988psychology,
  title     = {The Psychology of Everyday Things},
  author    = {Norman, Donald A.},
  year      = {1988},
  publisher = {Basic Books},
  address   = {New York, NY}
}
\clearpage
\appendix

\section{Notation}
\label{app:notate_table}
We define all formal notation used in earlier sections below.

\begin{table}[h]
\centering
\caption{Summary of notations.}
\begin{tabular}{ll}
\toprule
\textbf{Symbol} & \textbf{Description} \\
\midrule
$\mathcal{X}, \mathcal{T}, \mathcal{Z} \subset \mathbb{R}^d$ & Image, text, joint embedding space \\
$f_{\mathrm{img}}, f_{\mathrm{text}}$ & Image, text encoder\\
$x \in \mathcal{X}$ & Input image \\
$z_{img} \in \mathcal{Z}$ & Image embedding, $z_{img} = f_{\mathrm{img}}(x)$ \\
$t_a, t_{\bar{a}}$ & Text descriptions of affordance $a$ and its antonym \\
$z_a, z_{\bar{a}}$ & Text embeddings, $z_a = f_{\mathrm{text}}(t_a)$, $z_{\bar{a}} = f_{\mathrm{text}}(t_{\bar{a}})$ \\
$u_a$ & Affordance axis direction, $u_a = z_a - z_{\bar{a}}$ \\
$\lambda_a(z_{img})$ & Projection coordinate of $z_{img}$ onto the affordance axis \\
$\eta(x, a)$ & Calibration function estimates the probability that $a$ exists in $x$\\
$y, \hat{y}_a \in \{0,1\}$ & Ground-truth, predicted affordance label \\

$\mathcal{A}, \mathcal{A}_{\mathrm{base}}$ & Current and initial base affordance vocabularies \\
$\tau_i$ & Task context for image $x_i$ \\
$\mathcal{A}_i^{\mathrm{prop}}$ & Proposed affordance set for image $x_i$ \\
$\mathcal{A}_{i}^{\mathrm{base}}, \mathcal{A}_{i}^{\mathrm{new}}$ & Proposed base and newly discovered affordances for $x_i$ \\
$a_{ij} \in \mathcal{A}_i^{\mathrm{prop}}$ & $j$-th proposed affordance for image $x_i$ \\
$\tilde{y}_{ij} \in \{0,1\}$ & Generated label for pair $(x_i, a_{ij})$ \\
$\tilde{\mathcal{D}}_{\mathrm{disc}}$ & Automatically labeled discovery set \\

\bottomrule
\label{tab: notation}
\end{tabular}
\end{table}

% \paragraph{Encoder fine-tuning ablation.}
% The bottom block of Table~\ref{tab: baseline_performance_AE} ablates which encoder to fine-tune.
% On setting A, updating only the text encoder (frozen vision) is best at 96.2\%, slightly above updating both (94.2\%), while updating only the vision encoder (frozen text) is worst at 87.5\%.
% For in-distribution accuracy, the gain therefore comes mostly from moving the text-side affordance axes themselves in the embedding space, and the pretrained visual encoder can be reused as-is.
% On unseen object classes (setting C), the ordering flips: vision-only training is slightly ahead at 86.9\%.
% Vision-side updates begin to help once objects diverge from the training distribution.

\section{Per-Affordance Baseline Comparison}
\label{app:per-affordance}

\Cref{tab:all_baselines_affordance_results} provides a per-affordance accuracy breakdown for \ours{} and the four representative VLM baselines on an 80/20 image-level split on our dataset. Zero-shot VLM/embedding baselines perform substantially worse due to the lack of affordance-specific adaptation, while \ours{} achieves over 98\% average accuracy over base affordances and outperforms GPT-5.4 by 17\%. 
\begin{table}[h]
\centering
\caption{Per-affordance accuracy across all baselines on an 80/20 image-level split on our dataset.}
\label{tab:all_baselines_affordance_results}
\resizebox{\textwidth}{!}{%
\begin{tabular}{lcccccc}
\hline
\textbf{Affordance} & \textbf{Base CLIP} & \textbf{BLIP ITM} & \textbf{SigLIP} & \textbf{GPT-5.4} & \textbf{GPT-5.4 CoT} & \textbf{\ours{}} \\
\hline
Containable  & 0.667 & 0.538 & 0.538 & 0.795 & 0.744 & 1.000 \\
Graspable    & 0.538 & 0.564 & 0.564 & 0.821 & 0.872 & 0.974 \\
Liftable     & 0.462 & 0.205 & 0.513 & 0.718 & 0.718 & 0.974 \\
Movable      & 0.564 & 0.590 & 0.590 & 0.615 & 0.590 & 0.949 \\
Openable     & 0.436 & 0.385 & 0.308 & 0.923 & 0.897 & 1.000 \\
Rollable     & 0.205 & 0.436 & 0.641 & 0.949 & 0.872 & 1.000 \\
Stackable    & 0.538 & 0.487 & 0.385 & 0.897 & 0.821 & 0.974 \\
Supportable  & 0.564 & 0.436 & 0.641 & 0.769 & 0.744 & 0.974 \\
\hline
\textbf{Overall} & 0.497 & 0.455 & 0.522 & 0.811 & 0.782 & \textbf{0.981} \\
\hline
\end{tabular}%
}
\end{table}

\section{Incremental Affordance Learning}
\label{app:incremental}

\begin{figure}[!htbp]
    \centering
    \includegraphics[width=0.85\linewidth]{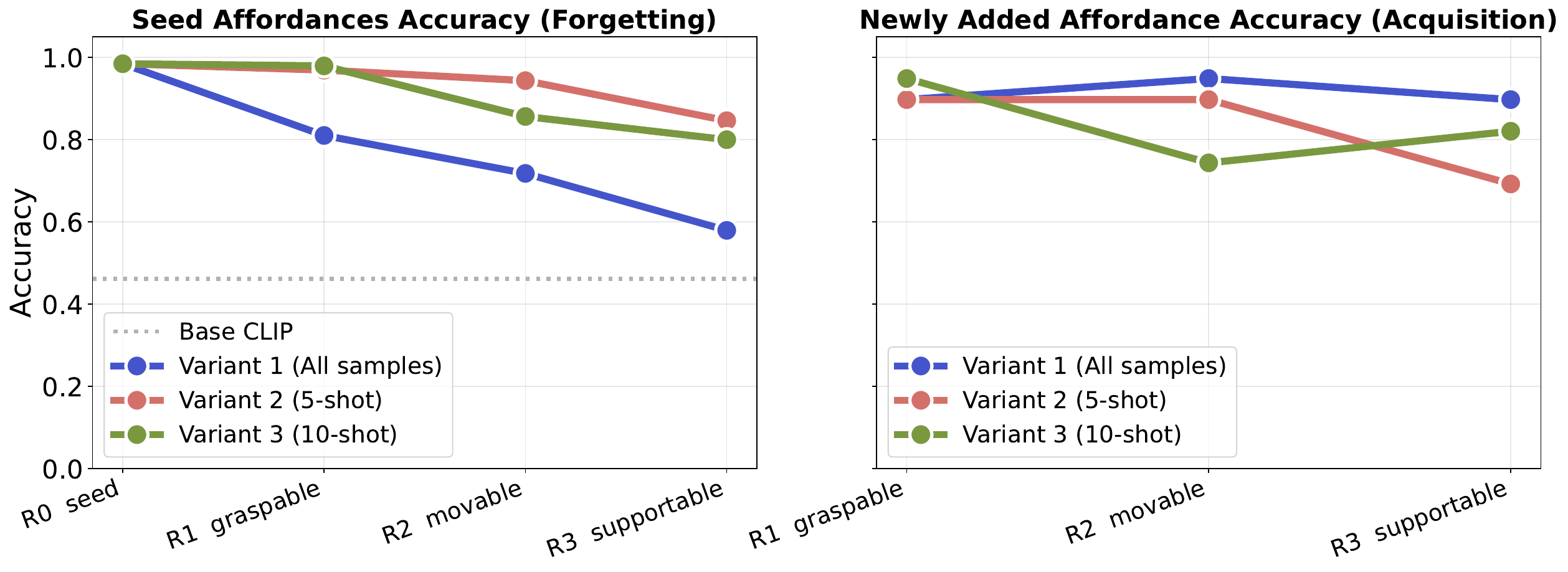}
    \caption{Incremental Learning: Seed-Affordance forgetting vs New-Affordance acquisition.}
    \label{fig:incremental_vars}
\end{figure}

We evaluate three incremental learning strategies (Fig.~\ref{fig:incremental_vars}) to study the stability–plasticity tradeoff when introducing new affordances without revisiting prior data.

\textbf{Variant 1 (all-samples training)} achieves strong acquisition of new affordances (up to 94.9\%), but suffers from severe catastrophic forgetting, with seed accuracy dropping from 98.5\% to 57.9\% after three rounds. This highlights the instability of naive continual fine-tuning.

\textbf{Few-shot variants (5-shot and 10-shot stratified sampling)} significantly mitigate forgetting. The 5-shot setting limits seed accuracy degradation to 13.8 percentage points, while the 10-shot setting provides a better balance between retention and acquisition. These results suggest that restricting updates to small, balanced datasets acts as an implicit regularizer, preserving seed accuracy.

% \begin{table}[!htbp]
% \centering
% \caption{Incremental Affordance Learning across three update strategies. All variants start from the same seed model (790 training images, 0.985 seed accuracy on liftable, containable, stackable, openable, rollable). Seed Acc.\ ($\Delta$) reports retention of the seed affordances after each round (raw accuracy and change from round 0); New Aff.\ Acc.\ is held-out test accuracy on the affordance added that round. Variant 1 trains on all 158 images per round and shows severe catastrophic forgetting (98.5\% $\to$ 57.9\%); 5-shot stratified preserves retention at the cost of new-affordance acquisition; 10-shot stratified balances the two.}
% \label{tab:incremental_variants}
% \resizebox{\textwidth}{!}{%
% \begin{tabular}{cl|cc|cc|cc}
% \hline
% & & \multicolumn{2}{c|}{\textbf{V1: All-Samples (158)}} & \multicolumn{2}{c|}{\textbf{V2: 5-Shot Strat.}} & \multicolumn{2}{c}{\textbf{V3: 10-Shot Strat.}} \\
% \textbf{Round} & \textbf{Added Affordance} & Seed Acc.\ ($\Delta$) & New Aff. & Seed Acc.\ ($\Delta$) & New Aff. & Seed Acc.\ ($\Delta$) & New Aff. \\
% \hline
% 0 & (seed)       & 0.985 (+0.000) & ---   & 0.985 (+0.000) & ---   & 0.985 (+0.000) & --- \\
% 1 & Graspable    & 0.810 ($-$0.174) & 0.897 & 0.969 ($-$0.015) & 0.897 & 0.979 ($-$0.005) & 0.949 \\
% 2 & Movable      & 0.718 ($-$0.267) & 0.949 & 0.944 ($-$0.041) & 0.897 & 0.856 ($-$0.128) & 0.744 \\
% 3 & Supportable  & 0.579 ($-$0.405) & 0.897 & 0.846 ($-$0.138) & 0.692 & 0.800 ($-$0.185) & 0.821 \\
% \hline
% \end{tabular}%
% }
% \end{table}

\section{Uncertainty-Guided VLM Fallback}
\label{app:uncertainty-fallback}

\begin{figure}[!htbp]
    \centering
    \includegraphics[width=0.85\linewidth]{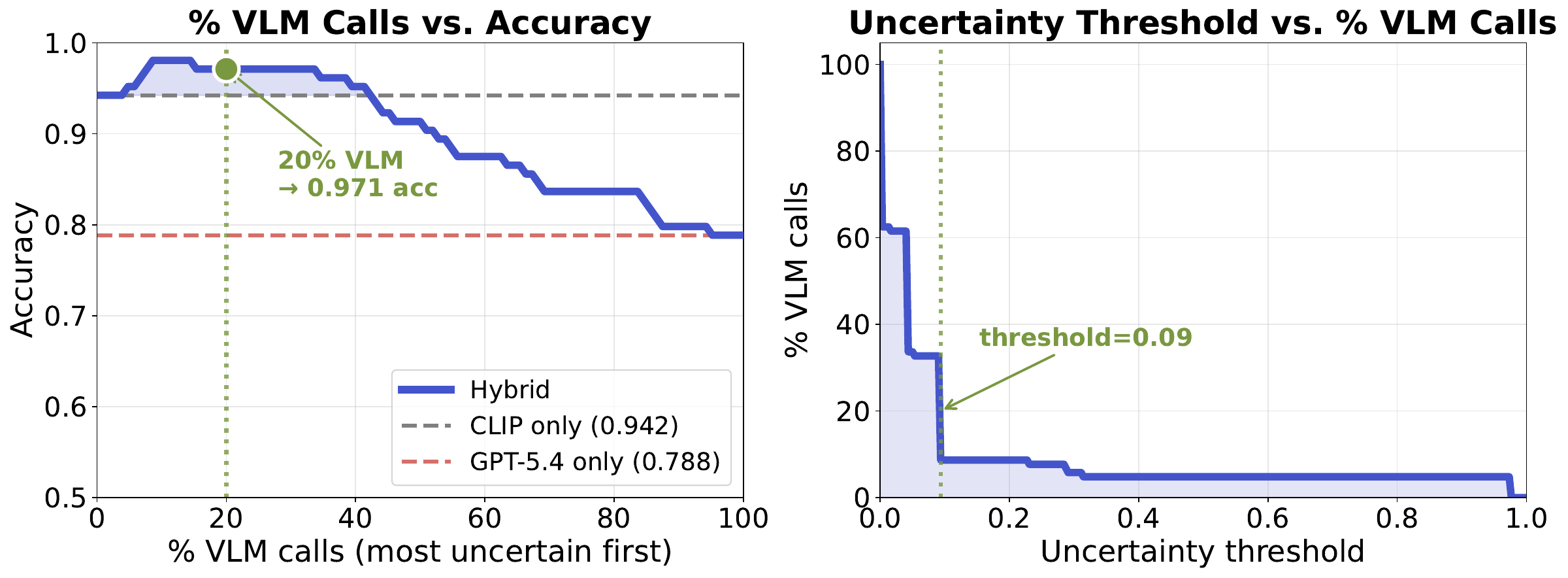}
    \caption{Accuracy vs Query Frequency tradeoff on seen image classes. }
    \label{fig:unc-threshold}
\end{figure}

We evaluate a setting in which GPT-5.4 serves as an oracle, invoked only when the uncertainty exceeds a threshold. Low-uncertainty predictions are handled directly by our framework while high-uncertainty cases receive VLM labeling. We extend the evaluation in \cref{fig:unc-accuracy} to seen object classes, achieving a peak 97.1\% accuracy while invoking VLM fallback for only 20\% of inferences. Fig.~\ref{fig:unc-threshold} sweeps the uncertainty threshold controlling this query frequency.
Overall, uncertainty serves as an effective trigger for selectively targeting likely failure cases among both seen and unseen object scenarios.

\section{Extended Related Work}
\label{app:extended_related}

\paragraph{Appearance-based affordance reasoning.}
Appearance-based visual affordance methods primarily focus on recognizing where actions can be applied in images or videos, often by predicting dense affordance maps over pixels or regions. Early work formulates affordance perception as a segmentation or detection problem, where each pixel or object proposal is assigned labels such as ``graspable'', ``pushable'', or ``supportable'', enabling downstream manipulation or navigation~\cite{hassanin2018visualaffordance,affordancenet,zeng2019learningvisualaffordances,zhu2023visualaffordancelearning,huang2023affordancesvideos,chuang2017learning2act,locate2023,iag2023,afformer2023}. While these models can achieve strong performance on curated datasets, they typically rely on task-specific supervision and category-centric architectures, which makes it difficult to transfer to new affordances or object types without collecting additional annotations and retraining. Moreover, because the representations are tightly coupled to visual appearance cues, they often fail to capture functional similarities between visually dissimilar objects that share the same affordance.

\paragraph{Functionality-based reasoning and interaction.}
A complementary line of work grounds affordances in robot interaction, learning which actions are feasible or useful directly from experience. Systems such as AffordanceNet and related frameworks learn to predict action-conditioned outcomes or value functions from interaction data, effectively capturing whether particular manipulations are likely to succeed~\cite{affordancenet,zeng2019learningvisualaffordances}. Other approaches, including PLATO and SayCan, combine structured reasoning with learned value estimates or language-guided policies, allowing robots to reason about possible sequences of actions in structured environments~\cite{belkhale2022plato,ahn2022saycan,wu2021imaginethat,mazzaglia2024informationaffordance,bahl2023hrp}. These methods excel when sufficient interaction data is available for the target platform and task distribution, but they tend to be specialized: affordances are defined with respect to a fixed robot embodiment, skill library, and set of tasks, and porting the system to new affordances or domains often requires new data collection and model retraining. In addition, many interaction-driven methods do not provide an explicit, low-dimensional representation of affordances that can be easily extended or inspected.

\paragraph{Generalization and latent affordance spaces.}
Beyond supervised and interaction-based approaches, several works explore learning continuous latent spaces that capture object functionality. These latent affordance spaces aim to encode how objects can be used, rather than what they look like, and can support tasks such as skill transfer or planning in novel scenarios~\cite{wu2021imaginethat,mazzaglia2024informationaffordance,huang2023affordancesvideos,bahl2023hrp}. However, when these spaces are learned purely from data without additional structure, they can be difficult to interpret or extend: adding a new affordance may require modifying the training objective or architecture, and the learned dimensions do not necessarily correspond to human-interpretable functional properties. This limits their usefulness as a modular interface for planning over affordances in open-world settings.

\paragraph{Vision--language models and semantic embeddings.}
More recent methods leverage semantic embedding spaces induced by vision--language models (VLMs) to relax closed-world assumptions and support open-vocabulary affordance reasoning. CLIP~\cite{radford2021learning} and related models align image and text encoders in a joint space where similarity corresponds to semantic compatibility, enabling zero-shot recognition and flexible prompting. Building on this, several works investigate using VLMs for affordance understanding, for example by matching object observations to textual affordance descriptors, generating affordance labels from natural language descriptions, or integrating VLMs into high-level planners~\cite{tang2024kalie,robopoint2024,birr2024autogptp,luo2022learningaffordance}. These approaches inherit the broad semantic coverage of pretrained VLMs, but practical use in robot planning remains challenging: prompted VLM queries are often slow, outputs may be coarse or lack spatial grounding, and the resulting representations are not always well-suited for uncertainty-aware control. Our framework builds on this line of work by structuring the CLIP embedding space into one-dimensional axes defined by affordance/antonym pairs, learning a calibrated mapping from positions on these axes to affordance probabilities, and integrating an affordance discovery mechanism that proposes and incorporates new axes when the existing memory is insufficient.

\section{Prompts for Affordance Generation and Labeling}
\label{app:prompts_aff_gen_labeling}

We provide the exact prompt templates used for affordance discovery and labeling. Each VLM call includes the cropped object image and a constructed prompt string, which may include the task context $\tau_i$ during discovery, and requests a JSON response using \texttt{response\_format=\{"type": "json\_object"\}}. The runtime variables used in the templates are summarized in Table~\ref{tab:prompt_runtime_variables}.

\begin{table}[h]
\centering
\caption{Runtime variables used in the affordance generation and labeling prompts.}
\label{tab:prompt_runtime_variables}
\resizebox{\textwidth}{!}{%
\begin{tabular}{ll}
\hline
\textbf{Variable} & \textbf{Description} \\
\hline
\texttt{ctx} & Per-image task context $\tau_i$ used during affordance discovery. \\
\texttt{mandatory} & Comma-separated base affordance memory $\mathcal{A}_{\mathrm{base}}$. \\
\texttt{ref\_extra} & Previously discovered non-base affordance names available for reuse. \\
\texttt{affordance} & Single affordance name queried during affordance labeling. \\
\hline
\end{tabular}%
}
\end{table}

\subsection{Shared Constants}
The following shared strings define the robot viewpoint used by both prompts and the naming convention used during affordance discovery.

\begin{lstlisting}[style=pythonprompt]
_VLM_ROBOT_VIEWPOINT = (
    "Perspective -- Judge affordances for a mobile manipulation robot (e.g. a Boston Dynamics "
    "Spot-class quadruped without an arm and gripper, or a robot arm with a gripper), not for a "
    "human. Each property describes whether the object supports that interaction for the robot "
    "given typical onboard sensing and end-effectors. The required base names are already "
    "defined in that spirit (e.g. graspable means the robot's gripper could meaningfully grasp it)."
)

_VLM_NAMING_CONSTRAINTS = (
    "Naming template -- every key (required or new) must match the same style as the eight base "
    f"robot affordances: {', '.join(BASE_VOCAB)}. Short snake_case property words in that same "
    "family (mostly -able / similar participles). Prefer ONE token; at most TWO only if both "
    "stay adjective-like (e.g. low_profile). No noun-phrase stacks.\n"
    "Do NOT output keys containing or starting with: has_, is_, with_, or ending in _present; "
    "do not use consistent_, accessible_, clearance, underseat_, pin_pull, or similar "
    "descriptor piles --  if unsure, omit the extra key.\n"
    "If a base affordance is already 1, do not add a longer compound that only repeats it "
    "(e.g. not handle_graspable when graspable is 1)."
)
\end{lstlisting}

\subsection{Affordance Generation}
\begin{lstlisting}[style=pythonprompt]
ctx = task.strip() or (
    "Infer robot-relevant affordances from the cropped object image alone."
)
prompt = (
    f"Task context: {ctx}\n\n"
    f"{_VLM_ROBOT_VIEWPOINT}\n\n"
    "Label a single cropped object for the robot. Return ONE JSON object: keys are snake_case "
    "affordance names; each value must be 0 or 1.\n"
    f"The REQUIRED keys list is the only naming template -- every other key (known extras or "
    f"new) must match that same word shape and brevity as those eight: [{mandatory}].\n"
    f"{_VLM_NAMING_CONSTRAINTS}\n\n"
    "REQUIRED -- include ALL of these keys (none missing), each with 0 or 1: "
    f"[{mandatory}]\n\n"
    "You MAY set 1 on additional known names when they apply (same object): "
    f"[{ref_extra}]\n\n"
    "ENCOURAGED -- add a few NEW keys only when a robot property is missing from the lists "
    "above. Each new key must be written in the SAME style as the eight REQUIRED keys "
    f"({mandatory}) -- not noun phrases, not a different convention. Use 1 only when "
    "visually grounded. Prefer fewer, cleaner extras."
)
\end{lstlisting}
\subsection{Affordance Labeling}
\begin{lstlisting}[style=pythonprompt]
prompt = (
    f"You are labeling one cropped object image for a single affordance property: \"{affordance}\".\n"
    f"{_VLM_ROBOT_VIEWPOINT}\n"
    "Do not use any task, instruction, or navigation context beyond the image and this property.\n"
    "Judge only the main object in the crop; ignore irrelevant background at the borders when possible.\n\n"
    "Reason step by step, but be terse. Fill these fields in order; later fields must be consistent "
    "with earlier ones:\n"
    "  1) main_object: short noun phrase for the dominant object you actually see in the crop "
    "(<= 8 words). If you cannot identify a single main object, say \"unclear\".\n"
    f"  2) definition: one-sentence operational meaning of \"{affordance}\" for this robot "
    "(what concrete physical interaction it implies).\n"
    "  3) evidence_for: 1 short sentence citing what visible features in the crop SUPPORT label 1. "
    "If none, write \"none\".\n"
    "  4) evidence_against: 1 short sentence citing what visible features SUPPORT label 0 "
    "(occlusion, wrong object class, scale, fixed/embedded, ambiguity). You MUST attempt this; "
    "if truly nothing, write \"none\".\n"
    "  5) confidence: one of \"high\", \"medium\", \"low\".\n"
    "     -- \"high\": the property is clearly supported by the visible main object -- a typical "
    "robot would succeed in the obvious way. Minor occlusion, partial crop, or background "
    "clutter are NOT reasons to drop below high if the relevant features are still visible.\n"
    "     - \"medium\": property is plausibly supported and you would lean toward 1, but there "
    "is a real caveat (e.g. shape is borderline, object is partially attached, scale is "
    "uncertain). Use this when you would say \"probably yes\".\n"
    "     - \"low\": main_object is \"unclear\", the property is clearly absent, or "
    "evidence_for and evidence_against are roughly balanced.\n"
    "  6) label: integer. Set 1 if confidence is \"high\" OR \"medium\"; set 0 if "
    "confidence is \"low\".\n\n"
    "Reply with exactly one JSON object, no markdown, with these keys: "
    "\"main_object\", \"definition\", \"evidence_for\", \"evidence_against\", \"confidence\", "
    "\"evidence\", \"label\". \"evidence\" should be a single short sentence (<=40 words) "
    "summarizing your decision; keep it for downstream readers.\n"
    "Example: {\"main_object\": \"red mug\", \"definition\": \"a parallel-jaw gripper can close "
    "around the object and hold it\", \"evidence_for\": \"thin handle and rim within gripper width\", "
    "\"evidence_against\": \"none\", \"confidence\": \"high\", "
    "\"evidence\": \"Mug handle gives a clean, small graspable feature.\", \"label\": 1}"
)
\end{lstlisting}

\subsection{Label Confidence Gate}
\begin{lstlisting}[style=pythonprompt]
if conf == "low":
    lab = 0
elif conf in ("high", "medium"):
    lab = raw_lab
else:
    lab = raw_lab
\end{lstlisting}
%===============================================================================

%===============================================================================

\clearpage
% The acknowledgments are automatically included only in the final and preprint versions of the paper.
% \acknowledgments{If a paper is accepted, the final camera-ready version will (and probably should) include acknowledgments. All acknowledgments go at the end of the paper, including thanks to reviewers who gave useful comments, to colleagues who contributed to the ideas, and to funding agencies and corporate sponsors that provided financial support.}

%===============================================================================

\end{document}